\documentclass[letterpaper, 10 pt, conference]{template}  

\IEEEoverridecommandlockouts                              

\usepackage[english]{babel}
\usepackage{mathtools}

\usepackage{amsmath}
\usepackage{graphicx}
\usepackage[colorlinks=true, allcolors=blue]{hyperref}
\usepackage{graphicx}
\usepackage{subcaption}
\usepackage{float}
\usepackage{amsfonts}  
\usepackage{placeins}
\usepackage{subcaption}
\usepackage{cite}
\usepackage[T1]{fontenc}

\DeclareCaptionFont{8pt}{\fontsize{8pt}{10pt}\selectfont}
\newcommand\scalemath[2]{\scalebox{#1}{\mbox{\ensuremath{\displaystyle #2}}}}

\title{Bearing-Only Tracking and Circumnavigation of a Fast Time-Varied Velocity Target Utilising an LSTM} 

\author{Mitchell Torok$^{1*}$, Mohammad Deghat$^{1}$, and Yang Song$^{2}$
\thanks{$^{*}$Corresponding Author: mitchell.torok@student.unsw.edu.au}
\thanks{$^{1}$ School of Mechanical and Manufacturing Engineering, University of New South Wales, Sydney, Australia, NSW, 2052}%
\thanks{$^{2}$ School of Computer Science and Engineering, University of New South Wales, Sydney, Australia, NSW, 2052}%
}

\begin{document}
\maketitle

\begin{abstract}
Bearing-only tracking, localisation, and circumnavigation is a problem in which a single or a group of agents attempts to track a target while circumnavigating it at a fixed distance using only bearing measurements. While previous studies have addressed scenarios involving stationary targets or those moving with an unknown constant velocity, the challenge of accurately tracking a target moving with a time-varying velocity remains open. This paper presents an approach utilising a Long Short-Term Memory (LSTM) based estimator for predicting the target's position and velocity. We also introduce a corresponding control strategy. When evaluated against previously proposed estimation and circumnavigation approaches, our approach demonstrates significantly lower control and estimation errors across various time-varying velocity scenarios. Additionally, we illustrate the effectiveness of the proposed method in tracking targets with a double integrator nonholonomic system dynamics that mimic real-world systems.
\end{abstract}


\section{Introduction}  

Target localisation and tracking is a problem in which a single or group of agents is tasked with estimating the position and following a target over time. This task becomes particularly complex when dealing with an uncooperative target whose state information is not directly accessible to the agent(s). In such scenarios, the agent(s) must rely on indirect observations to infer the target’s state. One employed method for tracking uncooperative targets involves the use of passive sensing systems. These systems collect bearing measurements, which provide the angle of the target relative to the agent. Sensors such as cameras or directional microphones collect this data, allowing the agent to determine the target’s direction without directly measuring its distance.

Bearing-only tracking is a well-explored problem, and in many cases, Kalman filter-based approaches have proven sufficient for target state estimation \cite{sindhu2019bearing}. However, recent research has also investigated deep learning methods to enhance performance in more complex scenarios \cite{zhao2021improved}. Moreover, the bearing-only tracking problem can vary depending on the number of available agents. For example, techniques have been proposed for multiple static agents collaborating to track targets \cite{wang2022passive,shalev2021botnet}. In contrast, a persistently exciting input trajectory is necessary to obtain an accurate estimation when using a single agent. In such cases, the agent may follow a specific motion pattern, such as a circular \cite{huang2021bearings} or sinusoidal \cite{ebrahimi2022bearing} trajectory, at an arbitrary location independent from the target.

The bearing-only target localisation and circumnavigation problem is an extension to bearing-only tracking, in which an agent aims to localise a target and circumnavigate it at a fixed radius. Previous work has explored this problem with a static target \cite{8483545,LI201818}, as well as a slowly moving target \cite{deghat2014localization,wang2022target}. The problem has also been extended to multi-agent scenarios \cite{MA20236712,sui2024collision}. These methods typically rely on estimators that iteratively update the target's approximate position based on successive bearing measurements. Recent work \cite{sui2024unbiased} introduced an unbiased estimator that ensures the target position and velocity estimation errors converge to zero when the target moves at an unknown constant velocity. However, these approaches are not directly targeted for cases where the target's velocity varies with time. A learning-based estimator could offer faster and more accurate target state estimation for the time-varied velocity case.

A Long Short-Term Memory (LSTM) \cite{hochreiter1997long} is a type of neural network architecture well-suited for learning patterns in sequential data over both short and long periods. LSTMs have demonstrated success in various tasks, such as time series forecasting and natural language processing \cite{van2020review}, where understanding context over time is crucial. The strength of an LSTM lies in its memory cell structure, which allows it to selectively retain or discard information as it processes a sequence of inputs. Previous research has explored the use of LSTMs for target-tracking applications. The work presented in \cite{zhu2024underwater} and \cite{patrick2024radar} both utilise LSTMs, incorporating both range and bearing measurements to enhance tracking accuracy. While these approaches show promise in successfully tracking targets, they stop before exploring the bearing-only case. In contrast, the work in \cite{cheng2022uuv} investigates a passive sonar sensor that produces bearing-only measurements and demonstrates the capabilities of bearing-only target tracking with a LSTM. However, in all of these works, the agent follows an arbitrary movement pattern that is not directly related to the target's motion.

\begin{figure}[htbp]
    \centering
    \includegraphics[width=0.85\columnwidth , trim=0cm 0.7cm 0cm 0.6cm, clip]{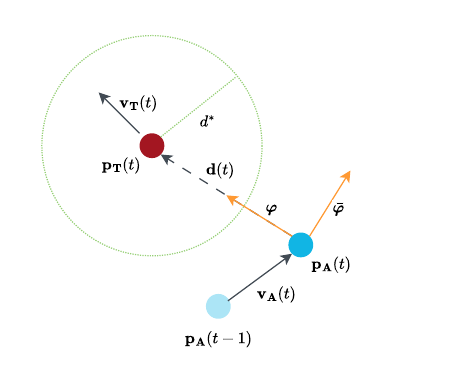}
    \caption{Geometric representation of an agent circumnavigating a moving target.}
    \label{fig:geo-rep}
\end{figure}

In this work, we present a bearing-only LSTM-based target estimator and an accompanying controller capable of achieving target localisation and circumnavigation for a target moving with time-varying velocity. The presented approach stores a history of previous bearing measurements and agent velocities, which are then inputted into an estimation model that predicts the target’s location and velocity at each timestep. This method enables the model to learn complex, time-varying velocity patterns, resulting in an agent that can track a target with reduced control and estimation error. Compared to existing LSTM-based target estimation approaches \cite{zhu2024underwater,patrick2024radar,cheng2022uuv}, the proposed approach is demonstrated for the circumnavigation case. Compared to bearing-only tracking and circumnavigation approaches \cite{deghat2014localization,wang2022target,sui2024unbiased}, our method is expressly suited for target trajectories with time-varied velocities. We demonstrate the performance of our system across a range of evaluation trajectories, highlighting its ability to achieve lower control and estimation errors in scenarios involving time-varying target velocities. An analysis of the proposed system's response to input noise is also presented.  

\section{Methodology}

\subsection{Controller}\label{sec:control}

Consider a target and an agent moving on an unconstrained 2D plane (Figure \ref{fig:geo-rep}), on which the agent is aiming to circumnavigate the target with a constant radius of $d^{*}$. Let the position of the target at time $t$ be defined as $\mathbf{p}_T(t) = [x_t(t),y_t(t)]^T \in \mathbb{R}^2$, and the velocity of the target as $\mathbf{v}_T(t) = \mathbf{\dot{p}}_T(t)  \in \mathbb{R}^2$. Similarly, let the position of the agent at time $t$ be defined as $\mathbf{p}_A(t) \in \mathbb{R}^2$, and the velocity of the agent as $\mathbf{v}_A(t) = \mathbf{\dot{p}}_A(t) \in \mathbb{R}^2$.

For each timestep, $t \ge 0$, a unit vector bearing measurement $\boldsymbol{\varphi}(t) \in \mathbb{R}^2$ pointing from the agent to the target is calculated using
\begin{equation}
 \boldsymbol{\varphi}(t)   = \frac{\mathbf{p}_T(t) - \mathbf{p}_A(t)}{ \| \mathbf{d}(t) \|},
\end{equation}
where $\mathbf{d}(t)\coloneqq \mathbf{p}_T(t) - \mathbf{p}_A(t) \in \mathbb{R}^2$ is a vector pointing from the agent to the target. A unit vector $\boldsymbol{\bar{\varphi}} \in \mathbb{R}^2$ is perpendicular to $\boldsymbol{\varphi}$ and is obtained by rotating  $\boldsymbol{\varphi}$ clockwise by $\pi/2$ radians, as shown in Figure \ref{fig:geo-rep}. Our approach stores the previous $l$ many bearing measurements and uses an LSTM model (explained in Section \ref{sec:LSTM}) to estimate the vector from the agent to the target $ \hat{\mathbf{d}}(t) $ and the velocity of the target, denoted by $\hat{\mathbf{v}}_T(t)$, at each timestep. We define the circumnavigation controller as:
\begin{equation}
\scalemath{0.90}{
\boldsymbol{u}(t) =
\begin{cases}
   k_t \boldsymbol{\bar{\varphi}}(t)& \text{if } t < l, \\
   k_t \boldsymbol{\bar{\varphi}}(t)+ k_r( \| \hat{\mathbf{d}}(t) \|  - d^{*})\boldsymbol{\varphi(t)} + \hat{\mathbf{v}}_T(t) & \text{if } t \ge l.
\end{cases}
}
\end{equation}
where $k_t$ and $k_r$ are the control gains affecting the agent's tangential and radial speed, respectively. Using the controller without the time-based condition could sometimes lead to unstable behavior, as the target estimator may produce inaccurate estimates when provided with a limited number of samples. Therefore, the distance error component of the controller is only activated after a sufficient number of samples have been collected.

\subsection{Target Estimation} \label{sec:LSTM}

\begin{figure}
    \centering
    \includegraphics[width=0.85\linewidth, trim=0cm 0cm 0cm 0cm, clip]{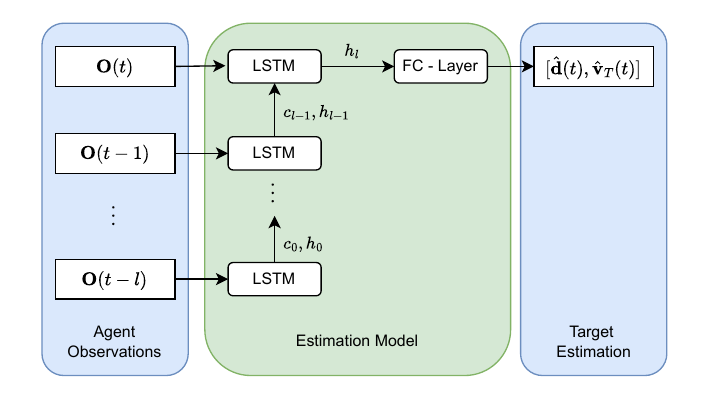}
    \caption{Target estimation model architecture. Here, $c$ denotes the cell state, which maintains long-term memory, while $h$ represents the hidden state, passed between each step in the LSTM layer.}
    \label{fig:Estimation_Arcitecture}
\end{figure}

Target estimation is carried out using a neural network model with an LSTM architecture. The architecture of an LSTM makes it well suited to learning temporal-spatial correlations. In this case, a many-to-one LSTM relationship is used, in which the LSTM receives a window of observations and only the final processing iteration is used in the model output. The proposed estimation model (Figure \ref{fig:Estimation_Arcitecture}) consists of an LSTM layer followed by a fully connected layer to map the output of the final LSTM iteration to the desired output size.

At each time step, an observation containing the bearing measurement of the target and the agent's velocity is generated, 
\begin{equation}
    \mathbf{O}(t) = [\boldsymbol{\varphi}(t),\mathbf{v}_A(t)].
\end{equation}
For a given LSTM window length $l$, the input to the LSTM model at time $t \ge l$ is given as 
\begin{equation}
    [\mathbf{O}(t), \mathbf{O}(t-1), \cdots , \mathbf{O}(t-l)].
\end{equation} 
The output of the LSTM is the estimated relative position and velocity of the target, 
\begin{equation}
    [\hat{\mathbf{d}}(t), \hat{\mathbf{v}}_T(t)]
\end{equation}
which is used by the controller (Section \ref{sec:control}).

\subsection{Training}
The estimation model and controller are inherently linked, as the estimation model’s output directly informs the controller, and the controller’s output results in the next input for the estimation model. Generating a ground truth dataset based on the assumption of a perfect estimation model and subsequently using supervised learning to train an estimation model for deployment with the controller resulted in poor system performance. This appeared to stem from the newly trained model's imperfect estimations, which led the controller to execute a new action, producing an agent trajectory outside the bounds of the ground truth dataset. Consequently, this created a feedback loop for the estimator, leading to further inaccuracies. This compounding effect frequently resulted in a loss of tracking, preventing the agent from correcting the radial distance error. To address this issue, an iterative method was employed to transfer target estimations gradually from the ground truth observer to the estimation model during the data collection process, thereby mitigating the feedback loop.

To train the target estimation model, an iterative on-policy training procedure was employed, where the agent's controller utilises either ground truth values, directly taken from the target state within the simulation, or the agent uses the output of the LSTM estimation model, to calculate the next control action during the dataset collection process. The training process occurs over many iterations to smoothly transition from an initial reliance on ground truth data to an end reliance on the LSTM estimation. At each training iteration, a new dataset is generated, followed by supervised learning using the newly collected dataset. Target trajectories for data collection during each iteration are generated by simulating the target patterns outlined in Section \ref{sec:eval}, with random initial conditions and system parameters, outside of the test parameter set, to create a diverse dataset. At each time step of the data generation phase, a random number $r \in [0, 1]$ is uniformly generated and compared against a switching threshold $s$ derived using the equation
\begin{equation}
s = - \frac{2}{I}i + 1,
\end{equation}
where $I$ is the total number of training iterations and $i$ is the current training iteration. In the case where $r < s $, the ground truth data $[\mathbf{d}(t), \mathbf{v}_T(t)]$ is used by the controller. In the case when $r \ge s $, the estimated data $[\hat{\mathbf{d}}(t), \hat{\mathbf{v}}_T(t)]$ is used by the controller. As training progresses, the resulting agent trajectories, and subsequently the dataset, become increasingly influenced by the estimation model, resulting in a smooth transition from the ground truth data to the estimation model.

For each iteration, a new dataset is generated, consisting of ground truth input-output pairs $[[\boldsymbol{\varphi}(t), \mathbf{v}_A(t)],[\mathbf{d}(t), \mathbf{v}_T(t)]]$, which are appended to the dataset at each timestep. As training occurs in simulation, ground truth information can be directly collected. Following this, supervised learning is applied to train the target estimation model. For a given LSTM input horizon length, the ground truth observations are input into the estimation model, and mean squared error is used to calculate a loss term based on the distance between the estimation model output $[\hat{\mathbf{d}}(t), \hat{\mathbf{v}}_T(t)]$ and the desired output $[\mathbf{d}(t), \mathbf{v}_T(t)]$.

\section{Evaluation} \label{sec:eval}

\subsection{Approach}

The supposed benefit of using an LSTM model for target estimation is that it increases the target-tracking capabilities of a circumnavigating agent compared to classical approaches. As such, factors like the agent's target estimation ability and circumnavigation performance with time-varied target trajectories are analysed and assessed. The proposed approach was tested against \cite{deghat2014localization} and \cite{sui2024unbiased}. These methods were implemented per their technical papers. These approaches do not use a learning-based approach to estimate the target's state, rather, they iteratively update it through successive bearing measurements. The approach presented in \cite{sui2024unbiased} improves on the method initially proposed in \cite{deghat2014localization} by introducing an unbiased estimator, enabling the system to converge to zero error when the target moves at an unknown yet constant velocity. No meaningful comparison could be made to other learning-based target state estimation approaches since they lack a target tracking or circumnavigation component. This evaluation specifically assesses tracking performance while attempting to maintain a fixed distance from the target.

All evaluated algorithms used the same control constants with a tangential gain of $k_t = 60$ and radial gain of $k_r = 10$, which were manually selected based on observation while testing. A constant circumnavigation radius of $d^{*} = 10$~m was set for the three algorithms. For all trials, the approach presented in \cite{deghat2014localization} used a position estimator gain of $12$, and the approach presented in \cite{sui2024unbiased} used a position estimator gain of $12$ and a velocity estimator gain of $6$. The parameters used in training for the LSTM estimator can be seen in Table \ref{tab:training_parameters}. The LSTM estimator model was trained once across all environments, and the same estimator model with fixed weights was used for all comparison trials. An operating frequency of $50$~Hz was used for all environments and algorithms. To assess the three algorithms, three target trajectory patterns were devised. For all experiments, the target started at the origin, $\mathbf{p}_T(0) = [0,0]$, and each algorithm started at $\mathbf{p}_A(0) = [15,0]$, with an initial target pose estimation of $\mathbf{\hat{p}}_T(0) = [5,0]$. All approaches were implemented in Python and ran concurrently for each trial.

Approaches are analysed based on their control error and estimation error. The control error is the error between the desired circumnavigation radius $d^{*}$ and the agent's current distance from the target $\|\mathbf{d}(t)\|$, formulated as,
\begin{equation}
\tilde{d}(t) = | \|\mathbf{d}(t)\| - d^{*} |.
\end{equation}
Likewise, the estimation error is the distance between the target location $\mathbf{p}_T(t)$ and the estimated location of the target $\hat{\mathbf{p}}_T(t)$, formulated as,
\begin{equation}
\tilde{p}_T(t) =  \| \mathbf{p}_T(t) - \hat{\mathbf{p}}_T(t)\|.
\end{equation}

\begin{table}[ht]

\centering
\scriptsize 
\setlength{\tabcolsep}{2pt} 
\renewcommand{\arraystretch}{1.4}
\caption{Parameters used for training the LSTM estimation model.}
\begin{tabular}{|c|c|}
\hline
 Parameter & Value  \\
\hline
Input size   & 4      \\
\hline
Output size   & 4      \\
\hline
Hidden size   & 512     \\
\hline
Horizon ($l$)   & 60     \\
\hline
Iterations ($I$)  & 50     \\
\hline
Iteration dataset size  & 100000     \\
\hline
Epochs   & 30     \\
\hline
Batch size   & 64    \\
\hline
Learning rate   & 0.001     \\
\hline
\end{tabular}
\label{tab:training_parameters}
\end{table}

\subsection{Constant velocity trajectory} \label{sec:constant_velocity}
\begin{figure}[h]
    \centering
    \begin{subfigure}{0.9\columnwidth}
        \centering
        \includegraphics[width=1.0\columnwidth]{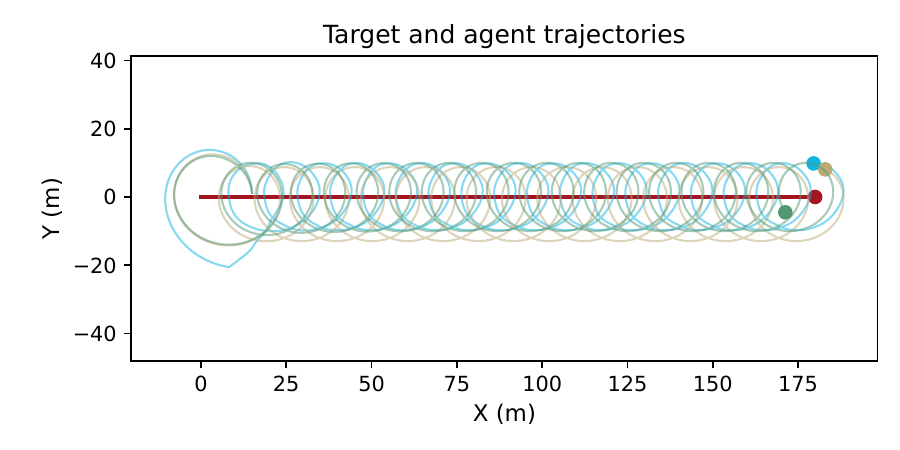}
    \end{subfigure}
    \begin{subfigure}{0.9\columnwidth}
        \centering
        \includegraphics[width=1.0\columnwidth]{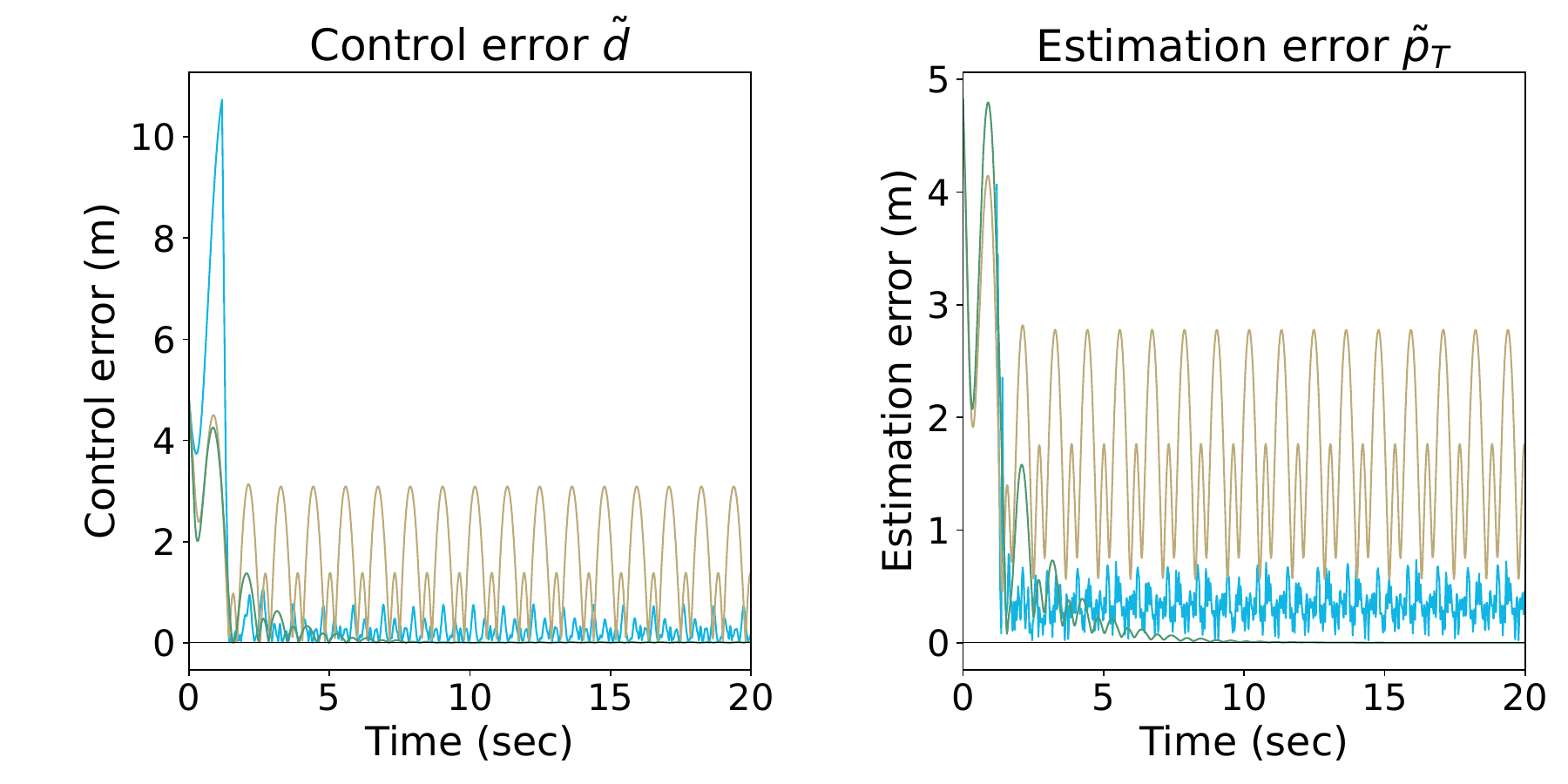}
    \end{subfigure}
    \begin{subfigure}{\columnwidth}
        \centering
        \includegraphics[width=1.0\columnwidth, trim=0cm 1.5cm 0cm 2.0cm, clip]{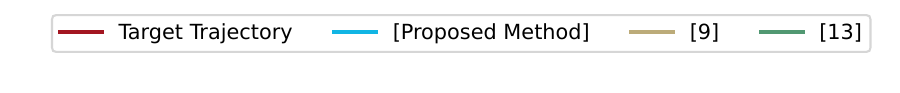}
    \end{subfigure}
    \caption{Result of the constant velocity simulation where $v = 9$~m/s. The agent markers indicate the final position of each respective system.}
    \label{fig:constant_vel}
\end{figure}

\begin{figure}[h]
    \centering
    \begin{subfigure}{0.9\columnwidth}
        \centering
        \includegraphics[width=1.0\columnwidth]{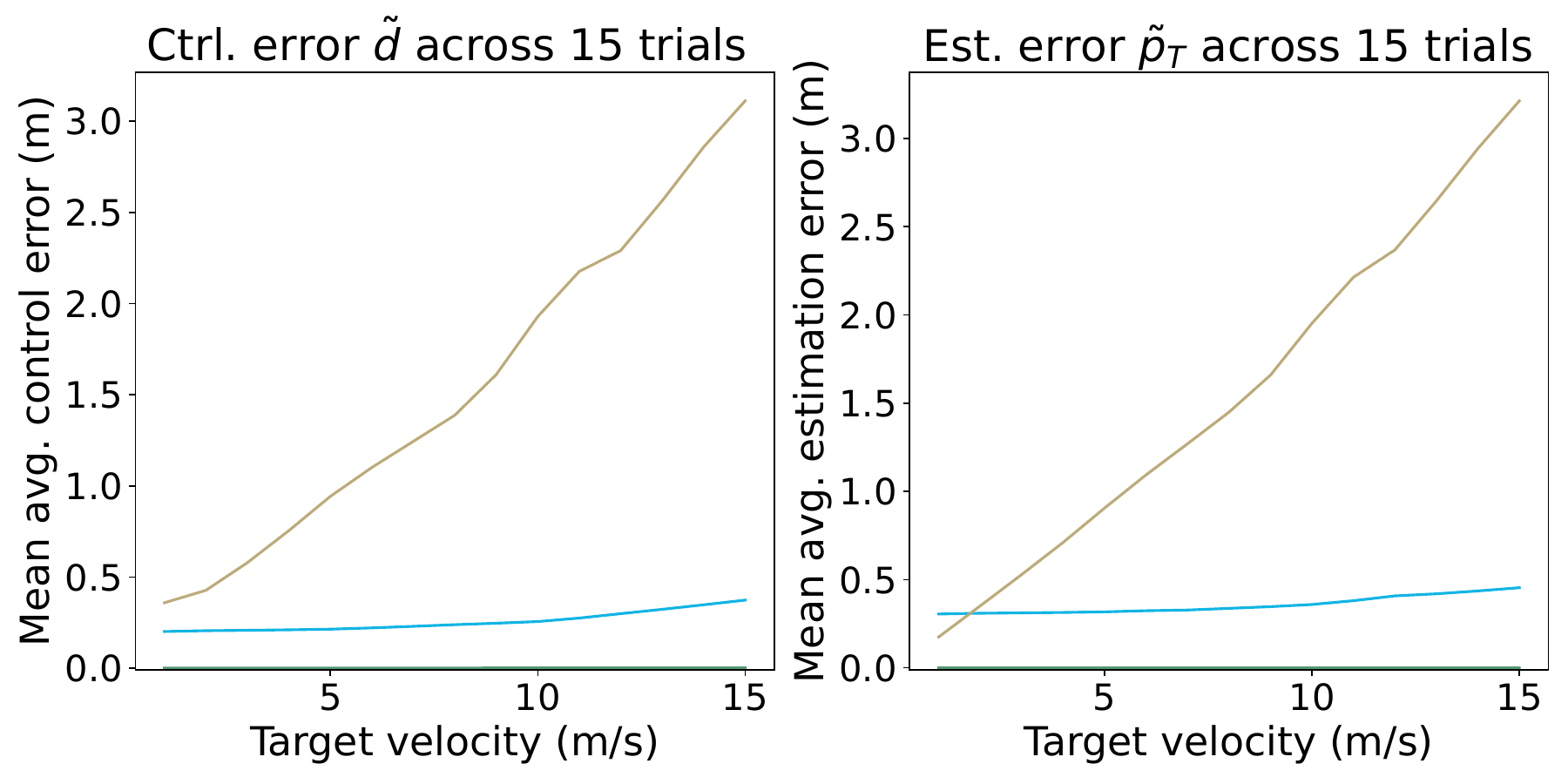}
    \end{subfigure}
    \begin{subfigure}{\columnwidth}
        \centering
        \includegraphics[width=1.0\columnwidth, trim=0cm 2cm 0cm 2cm, clip]{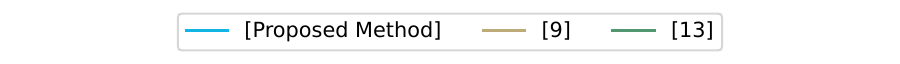}
    \end{subfigure}
    \caption{Comparison of each agent's control and estimation error across fifteen simulations each using a constant velocity for $1000$ timesteps. For each trial, only the values from the last $5$ seconds were averaged, allowing the initial $15$ seconds for each system to settle.}
    \label{fig:constant_vel_eval}
\end{figure}

The constant velocity case evaluates an environment where the target moves at a fixed velocity, which is unknown to the agent (Figure \ref{fig:constant_vel}). The target follows the dynamic system:
\begin{equation}
\begin{aligned}
\dot{x_t} &= v \\
\dot{y_t} &= 0,
\end{aligned}
\end{equation}
where \(v\) represents the target's linear velocity. To assess the performance of each algorithm, fifteen simulations were conducted, each lasting $1000$ timesteps, with the target maintaining a constant velocity \(v\) throughout the trial. Each simulation used a different velocity that was evenly selected between $1$ and $15$~m/s (Figure \ref{fig:constant_vel_eval}). For the evaluated cases, all the algorithms successfully managed to track the target with low error. Notably, \cite{sui2024unbiased} is demonstrated to converge to zero estimation error across all constant velocity trials.

\subsection{Circle trajectory} \label{sec:circle}
\begin{figure}[h]
    \centering
    \begin{subfigure}{0.9\columnwidth}
        \centering
        \includegraphics[width=1.0\columnwidth]{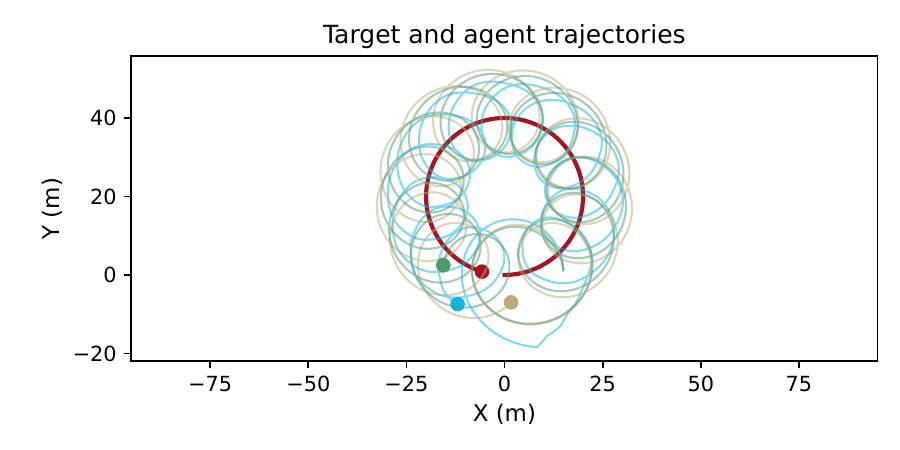}
    \end{subfigure}
    \begin{subfigure}{0.9\columnwidth}
        \centering
        \includegraphics[width=1.0\columnwidth]{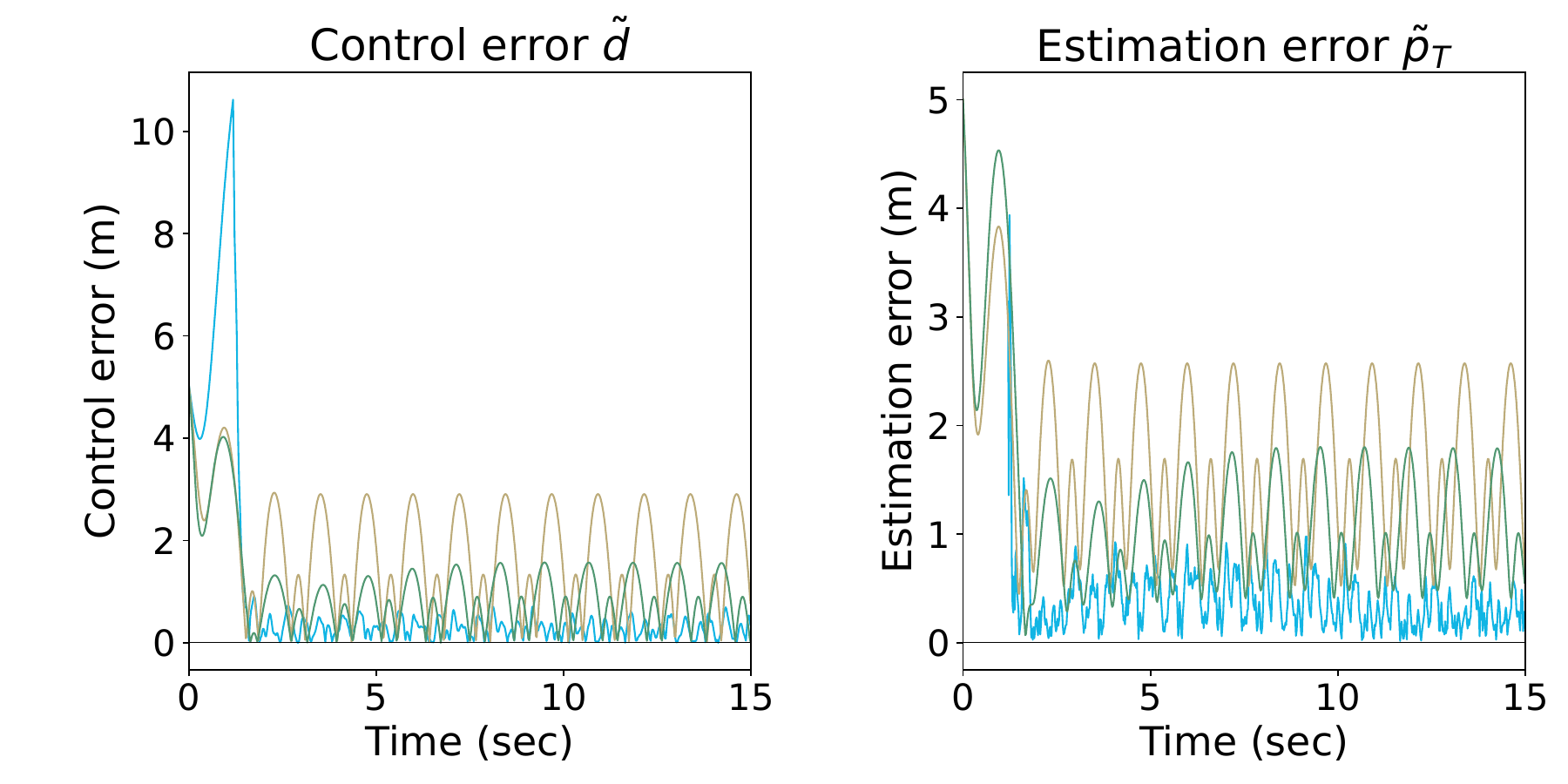}
    \end{subfigure}
    \begin{subfigure}{\columnwidth}
        \centering
        \includegraphics[width=1.0\columnwidth, trim=0cm 1.5cm 0cm 2.0cm, clip]{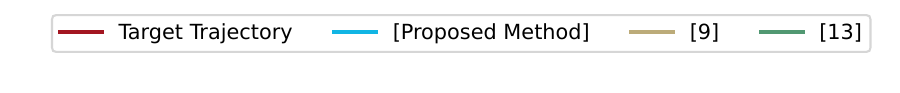}
    \end{subfigure}
    \caption{Result of the circle tracjetory simulation where $ \omega = 0.4$~rad/s . The agent markers indicate the final position of each respective system.}
    \label{fig:circle}
\end{figure}

\begin{figure}[h]
    \centering
    \begin{subfigure}{0.9\columnwidth}
        \centering
        \includegraphics[width=1.0\columnwidth]{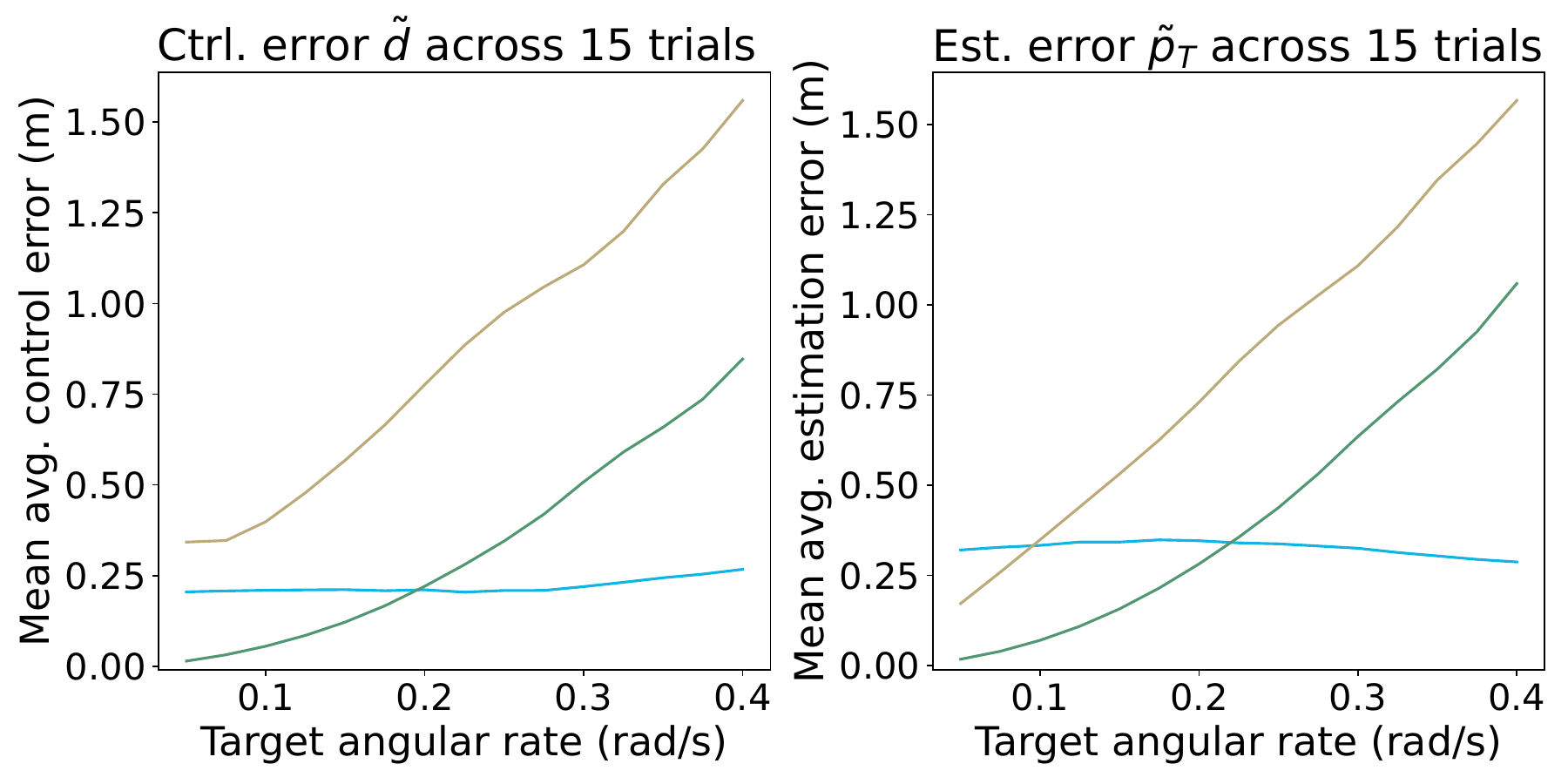}
    \end{subfigure}
    \begin{subfigure}{\columnwidth}
        \centering
        \includegraphics[width=1.0\columnwidth, trim=0cm 2cm 0cm 2cm, clip]{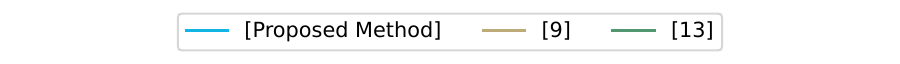}
    \end{subfigure}
    \caption{Comparison of each agent's control and estimation error across fifteen simulations of the circle trajectory case. Each trial used a constant angular velocity for $750$ time steps. For each trial, only the values from the last $5$ seconds were averaged, allowing the initial $10$ seconds for each system to settle.}
    \label{fig:circle_eval}
\end{figure}

Unlike a constant velocity case, the circle case moves the target in a circular pattern, allowing for a time-varied velocity (Figure \ref{fig:circle}). The trajectory of the target is defined by the dynamics:
\begin{equation}
\begin{aligned}
x_t(t) &= r\cos( \omega t - \pi/2)\\
y_t(t) &= r\sin( \omega t - \pi/2) + r , \\
\end{aligned}
\end{equation}
where $\omega$ is the constant angular rate and $r$ is the fixed target radius. For this experiment, the radius was fixed at $r = 20$~m, and the angular rate of the target was varied between $0.05$ to $ 0.4$~rad/s. Fifteen simulations, each lasting $750$ timesteps, were conducted, each with evenly spaced angular rate parameters (Figure \ref{fig:circle_eval}). During each trial, the angular rate remained constant. The three algorithms tracked the target at a low target angular rate with little error. However, as the speed of the target increased, the performance of the approaches presented in \cite{deghat2014localization} and \cite{sui2024unbiased} decreased notably, while the proposed approach maintained consistent performance.

\begin{table}[htbp]
\vspace*{0.3cm}
\centering
\scriptsize
\setlength{\tabcolsep}{1pt} 
\renewcommand{\arraystretch}{1.4}

\caption{Comparison of circumnavigation tracking approaches for the double integrator nonholonomic target movement case. A total of $1000$ trials, each $1000$ timesteps in length, were conducted with the tracking and control errors recorded for all three algorithms. Each entry shows the mean average and the associated standard deviation. }
\label{tab:}

\begin{tabular}{|c|c|c|}
\hline
   & \multicolumn{2}{|c|}{Double integrator nonholonomic target}\\
\hline
Method & Control error (m) & Estimation error (m) \\
\hline
\cite{deghat2014localization}  &  2.406 $\pm$ 1.440 & 2.401 $\pm$ 1.516 \\
\hline
\cite{sui2024unbiased}  & 1.658 $\pm$ 1.092 & 1.531 $\pm$ 1.128\\
\hline
[Proposed Method]  & \textbf{0.819 $\pm$ 0.236} & \textbf{0.524  $\pm$ 0.101}  \\

\hline
\end{tabular}
\end{table}

\subsection{Nonholonomic trajectory} \label{sec:non_holonomic}

\begin{figure}[h]
    \centering
    \begin{subfigure}{0.9\columnwidth}
        \centering
        \includegraphics[width=1.0\columnwidth]{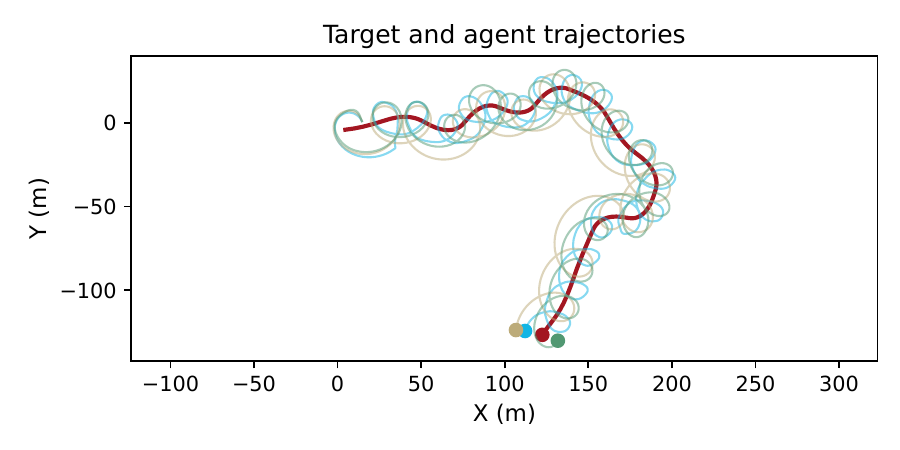}
    \end{subfigure}

    \begin{subfigure}{\columnwidth}
        \centering
        \includegraphics[width=1.0\columnwidth, trim=0cm 1.5cm 0cm 1.5cm, clip]{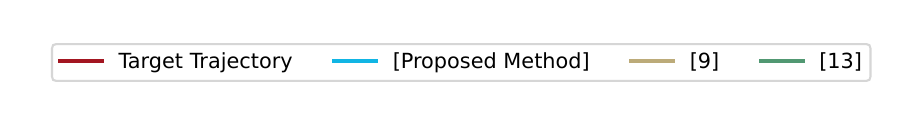}
    \end{subfigure}
    \caption{Example of target and agent trajectories for the nonholonomic target trial. The agent markers indicate the final position of each respective system.}
    \label{fig:noholonomic_example}
\end{figure}

The nonholonomic environment aims to replicate how a car or boat may move, as such, system parameters were constrained to mimic real-world systems. A double integrator nonholonomic approach was used (Figure \ref{fig:noholonomic_example}). To generate a random and continuously changing target trajectory, every $75$ timesteps, new input parameters were randomly selected between the parameter limits with a uniform distribution.

The nonholonomic case follows the dynamic equations:
\begin{equation}
\begin{aligned}
\dot{v_t} &= a \\
\dot{\omega_t} &= \alpha \\
\dot{x_t} &= v_t \cos(\theta_t) \\
\dot{y_t} &= v_t \sin(\theta_t) \\
\dot{\theta_t} &= \omega_t,
\end{aligned}
\end{equation}
where $a$ is the magnitude of the linear acceleration and $\alpha$ is the magnitude of the angular acceleration. The velocity parameters were constrained as $0$~m/s $ \leq v_t \leq 20$~m/s and $-\pi/2 $~rad/s $ \leq \omega_t \leq \pi/2 $~rad/s. The acceleration parameters were constrained as $-5 \text{~m/s}^2 \leq a \leq 5 \text{~m/s}^2$ and $-\pi/2 \text{~rad/s}^2 \leq \alpha \leq \pi/2 \text{~rad/s}^2$. When a new selection of system parameters occurs, only $a$ and $\alpha$ are randomly selected from a uniform distribution. Additionally, the angular velocity of the target $\omega_t$ is reset to zero to prevent trajectories in which the target would continually spiral for the entire trial.

\begin{figure}[h]
    \centering
    \begin{subfigure}{0.9\columnwidth}
        \centering
        \includegraphics[width=1.0\columnwidth]{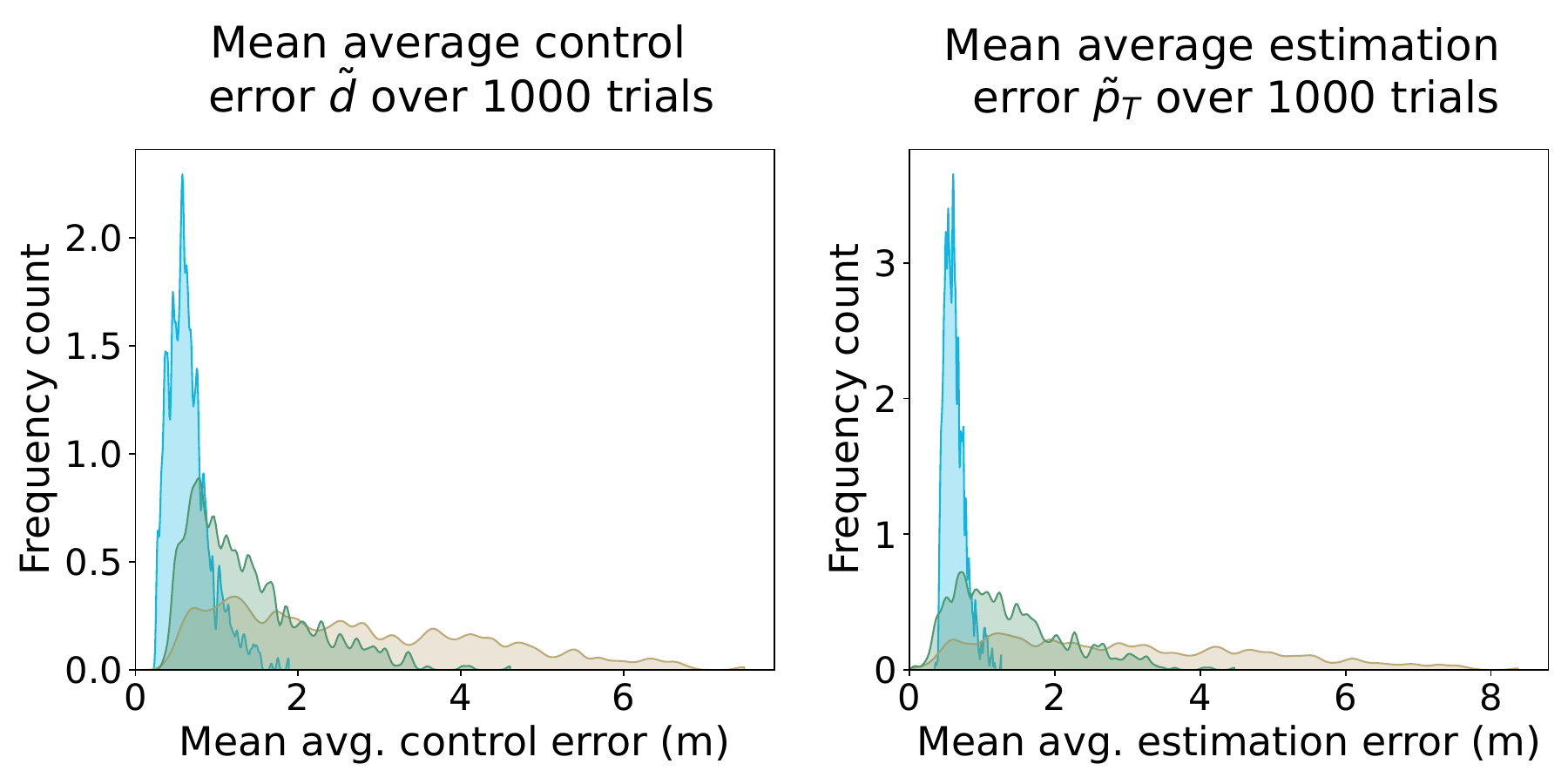}
    \end{subfigure}
    \begin{subfigure}{\columnwidth}
        \centering
        \includegraphics[width=1.0\columnwidth, trim=0cm 2cm 0cm 2cm, clip]{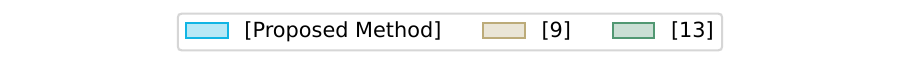}
    \end{subfigure}
    \caption{A histogram depicting the performance of the three algorithms over $1000$ trials, conducted while tracking the double integrator nonholonomic target trajectory.}
    \label{fig:double_int_hist}
\end{figure}

To evaluate the algorithms, $1000$ trials, each spanning $1000$ timesteps, were conducted for the double integrator nonholonomic case. After each trial, the mean average tracking error and estimation error were calculated. The performance metrics of each agent were averaged across the simulations. Table \ref{tab:} summarises the mean and standard deviation of the control and estimation errors across these trials. Figure \ref{fig:double_int_hist} presents histograms detailing the control and estimation error over the trails. The histograms indicate that the proposed method yields a lower mean error and exhibits a narrower error distribution compared to the other algorithms.

\subsection{Input noise}

\begin{figure}[htbp]
    \centering
    \begin{subfigure}{0.9\columnwidth}
        \centering
        \includegraphics[width=1.0\columnwidth]{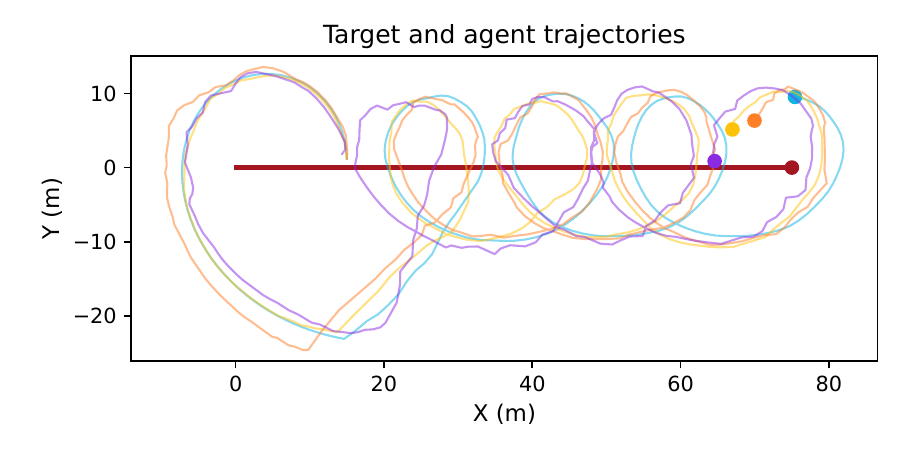}
    \end{subfigure}
    \begin{subfigure}{\columnwidth}
        \centering
        \includegraphics[width=1.0\columnwidth, trim=0cm 0cm 0cm 0cm, clip]{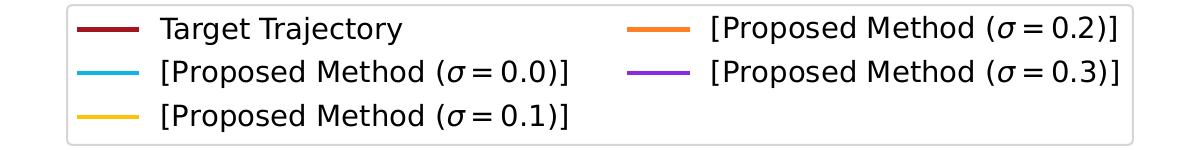}
    \end{subfigure}
    \caption{Visualisation of the proposed method with varying levels of input noise as it tracks a target moving at a constant velocity of $v = 15$ m/s, over $250$ timesteps. }
    \label{fig:noise_example}
\end{figure}

\begin{table}[htbp]
\vspace*{0.3cm}
\centering
\scriptsize 
\setlength{\tabcolsep}{1pt} 
\renewcommand{\arraystretch}{1.4}

\caption{Comparison of the proposed method when the inputs are perturbed by normally distributed random noise. For each noise reading, a new estimation model was trained with the perturbed noise applied to all environments during data collection. A total of $1000$ trials, each $1000$ timesteps long, were conducted with the tracking and error recorded for the evaluated estimation models. Each entry shows the mean average and the associated standard deviation.}
\label{tab:noise}

\begin{tabular}{|c|c|c|}
\hline
   & \multicolumn{2}{|c|}{Double integrator nonholonomic target}\\
\hline
Method & Control error (m) & Estimation error (m) \\
\hline
[Proposed Method ($\sigma = 0.0$)]  & 0.843 $\pm$ 0.282 & 0.628 $\pm$ 0.282   \\
\hline
[Proposed Method ($\sigma = 0.1$)]  & 1.091 $\pm$ 0.327 & 1.091 $\pm$ 0.327   \\
\hline
[Proposed Method ($\sigma = 0.2$)]  & 1.161 $\pm$ 0.289 & 1.161 $\pm$ 0.289  \\
\hline
[Proposed Method ($\sigma = 0.3$)]  & 1.414 $\pm$ 0.341 & 1.414 $\pm$ 0.341  \\
\hline
\end{tabular}
\end{table}

\begin{figure}[htbp]
    \centering
    \begin{subfigure}{0.9\columnwidth}
        \centering
        \includegraphics[width=1.0\columnwidth]{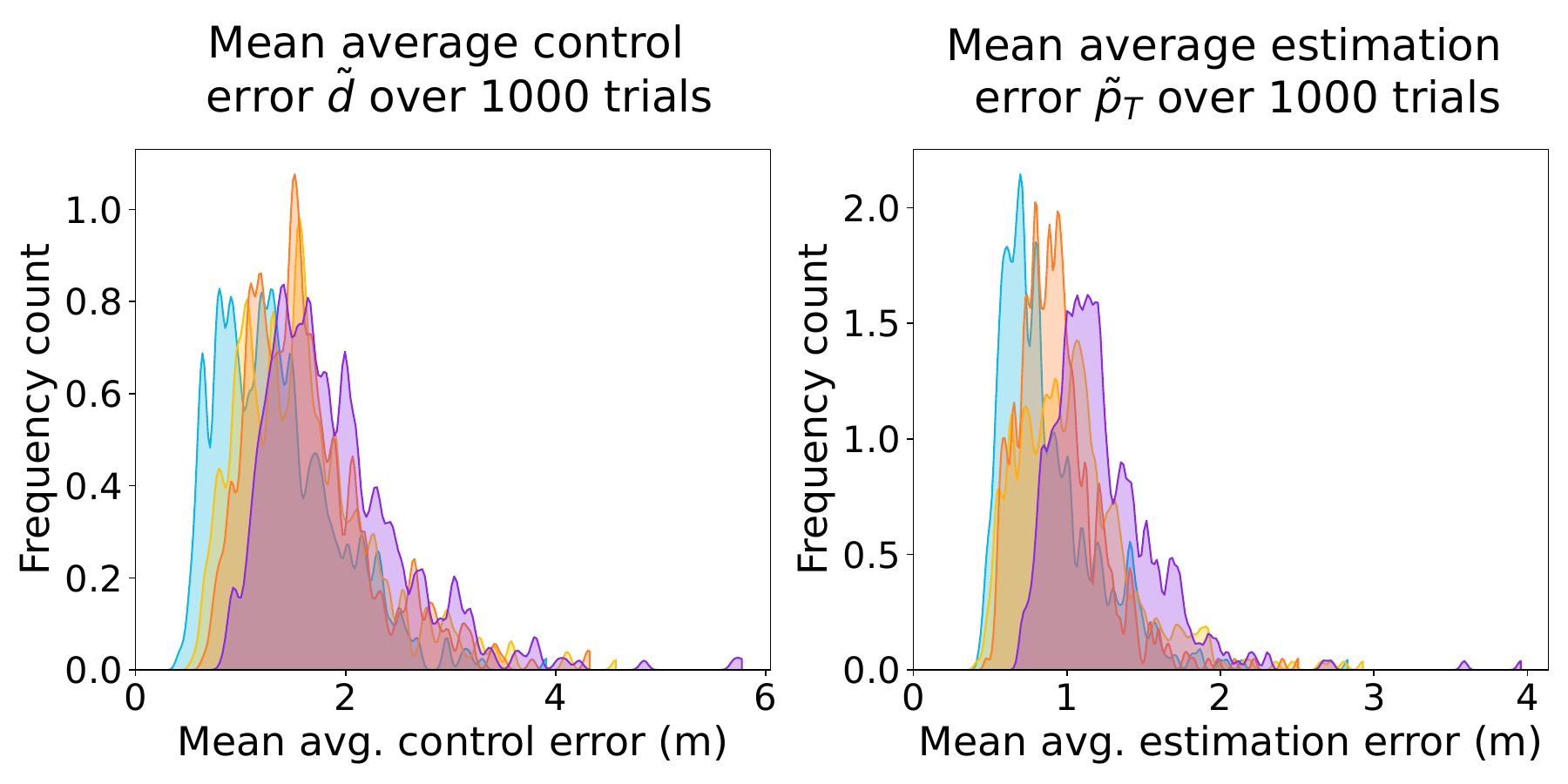}
    \end{subfigure}
    \begin{subfigure}{\columnwidth}
        \centering
        \includegraphics[width=1.0\columnwidth, trim=0cm 1.5cm 0cm 1.5cm, clip]{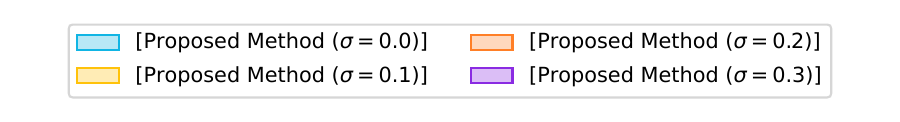}
    \end{subfigure}
    \caption{A histogram depicting the performance of the proposed method when trained and evaluated on different levels of input noise.}
    \label{fig:noise_eval}
\end{figure}

\begin{figure*}[htbp]
    \centering
    \begin{subfigure}{0.24\textwidth}
        \centering
        \includegraphics[width=\textwidth]{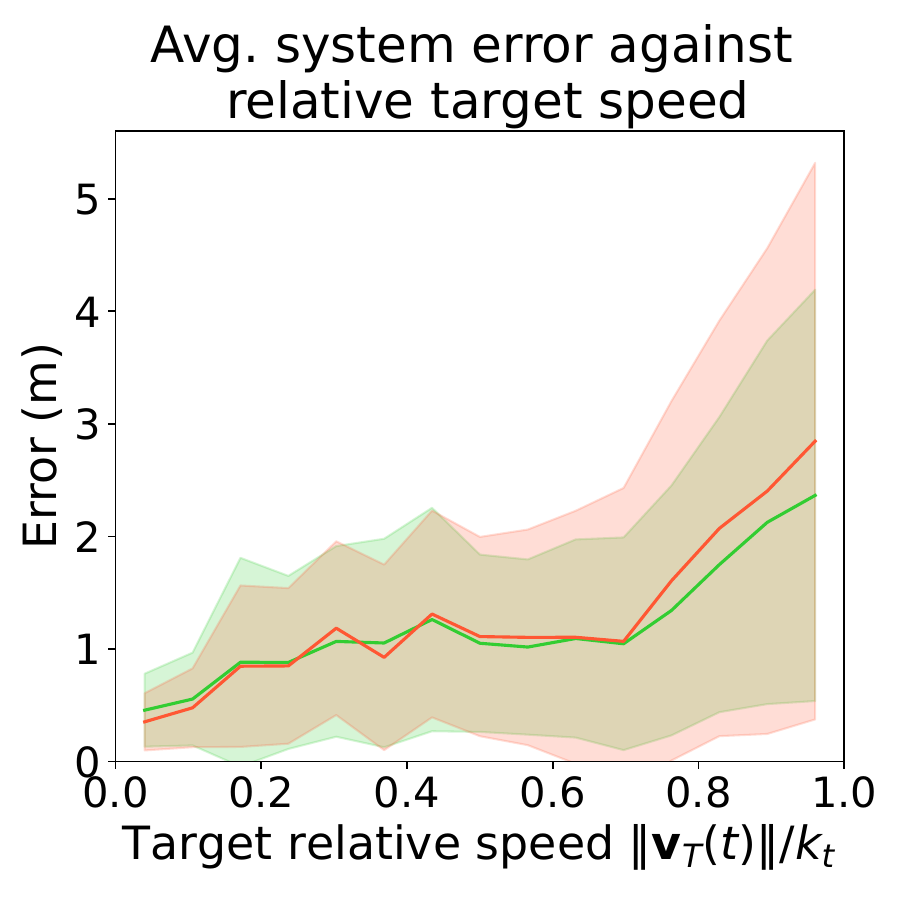}
        \caption{Constant velocity case}
    \end{subfigure}
    \hfill
    \begin{subfigure}{0.24\textwidth}
        \centering
        \includegraphics[width=\textwidth]{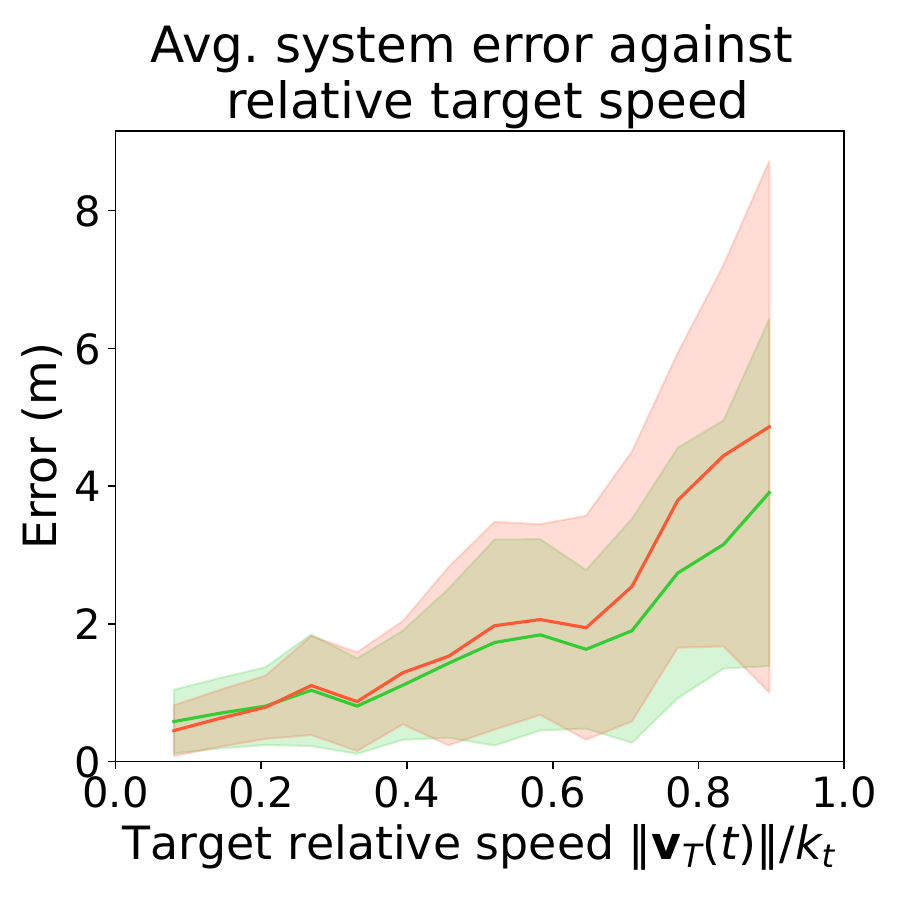}
        \caption{Circle case}
    \end{subfigure}
    \hfill
    \begin{subfigure}{0.24\textwidth}
        \centering
        \includegraphics[width=\textwidth]{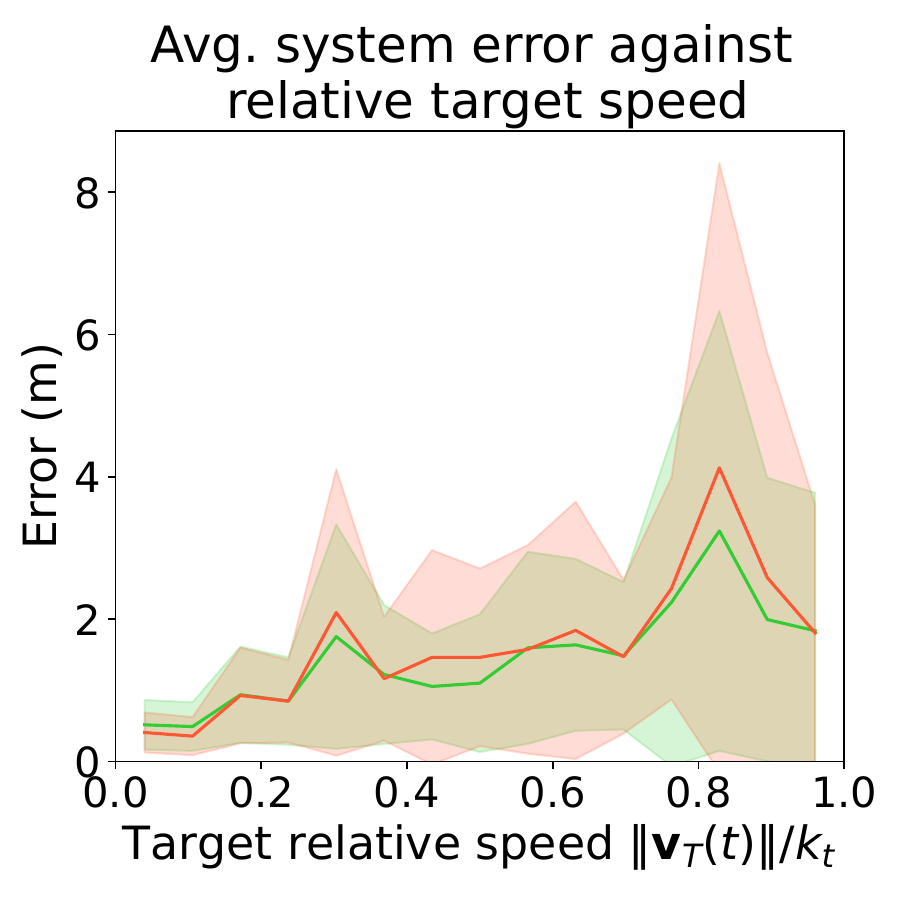}
        \caption{Double integrator case}
    \end{subfigure}

    \begin{subfigure}{\columnwidth}
        \centering
        \includegraphics[width=1.0\columnwidth, trim=0cm 0.5cm 0cm 0cm, clip]{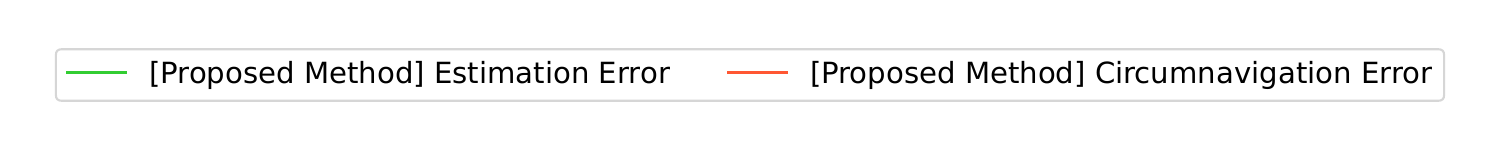}
    \end{subfigure}
    
    \caption{Performance graphs illustrating the system's tracking ability for a fast-moving target. Four target trajectory cases are presented, each with 15 trials of 500 time steps conducted at evenly spaced maximum speeds. The mean errors and standard deviations are shown relative to the target's speed compared to the agent’s tangential speed.}

    \label{fig:fast_test}
\end{figure*}

\begin{figure}[htbp]
    \centering
    \begin{subfigure}{0.9\columnwidth}
        \centering
        \includegraphics[width=1.0\columnwidth]{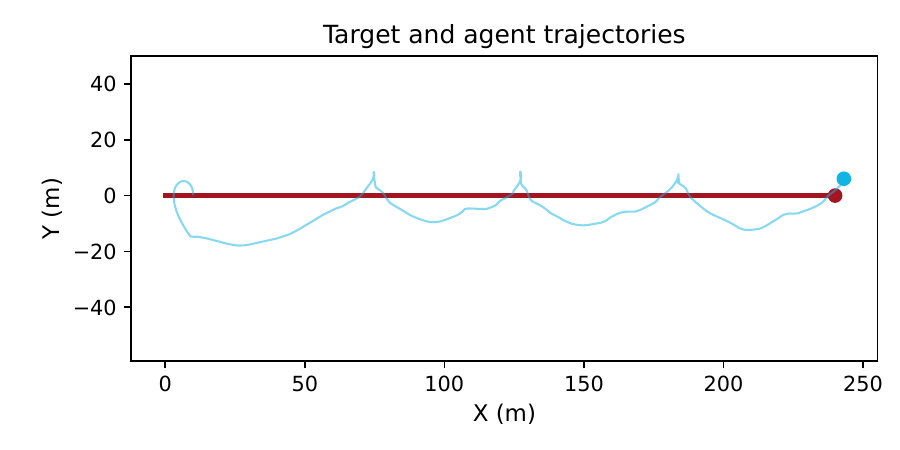}
    \end{subfigure}
    \begin{subfigure}{\columnwidth}
        \centering
        \includegraphics[width=1.0\columnwidth, trim=0cm 2cm 0cm 2cm, clip]{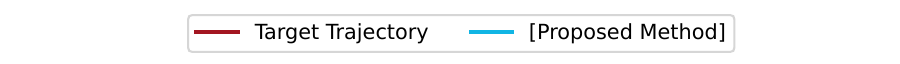}
    \end{subfigure}
    \caption{Tracking performance of the proposed method following a target moving at a constant velocity, with a speed ratio of $\| \boldsymbol{v}_T(t) \| / k_t = 0.96$. The target moves at a velocity of $24$~m/s, while the agent maintains a tangential speed of $25$~m/s for $500$ timesteps.}
    \label{fig:fast_demo}
\end{figure}

When deploying this approach in a real-world system, some quantity of sensor input noise may be expected. In such cases, estimation models can be trained with the expected noise applied to the inputs during the data collection phase. To evaluate this, new estimation models were developed, each incorporating a different level of noise in the input measurements (Figure \ref{fig:noise_example}). A normally distributed noise with a mean of zero and fixed standard deviation was applied to all observations during data collection, affecting both the bearing measurement, \(\boldsymbol{\varphi}(t)\), and the agent's velocity, \(\mathbf{v}_A(t)\). Four estimation models were assessed, corresponding to noise standard deviations of \(\sigma = 0.0\), \(0.1\), \(0.2\), and \(0.3\).

For each estimation model, $500$ trials of $500$ time steps were conducted using the double integrator nonholonomic trajectory. All models were evaluated with the same target trajectories and initial conditions, however, each model ran independently with noise applied according to the specified standard deviation for that estimation model. The results of these trials can be seen in Table \ref{tab:noise} and visualised as a histogram in Figure \ref{fig:noise_eval}. These results demonstrate that, while input noise slightly reduces performance, the estimation model successfully tracked the target even with a noise standard deviation of $\sigma = 0.3$, albeit producing a noticeably noisy agent trajectory.

\subsection{Fast target}
This experiment investigates the performance of the proposed method when tracking a fast-moving target, defined as a scenario in which the target's speed approaches the agent's tangential speed, 
\begin{equation}
    \frac{\|\boldsymbol{v}_T(t)\|}{k_t} \to 1.
\end{equation}
The fast target scenario represents an extreme case, as circumnavigation tasks typically assume that the agent moves significantly faster than the target. As shown in Figure~\ref{fig:fast_demo}, the agent’s trajectory here is visibly distinct from the previous constant velocity trials (Figure~\ref{fig:constant_vel}). Nonetheless, this scenario may arise if the agent’s tangential speed is physically constrained and a sufficiently fast target approaches this limit. Thus, examining this fast-moving target case demonstrates the ability of the proposed estimation model and controller in such cases.

To assess the proposed method’s ability to track fast-moving targets, the onboard estimation model was retrained with a tangential gain of $k_t = 25$ and radial gain of $k_r = 4$, better matching realistic deployment speeds. Testing included constant velocity, circular, and double integrator nonholonomic trajectories. In the constant velocity case, the target speed ranged from $1$ to $24$~m/s. For the circular trajectory, angular rates were set between $0.1$ and $1.2$~rad/s. In the nonholonomic case, the target speed was fixed per trial, with yaw rates chosen randomly. All other parameters followed the setups in Sections~\ref{sec:constant_velocity}, \ref{sec:circle}, and \ref{sec:non_holonomic}.

For each environment, fifteen trials were conducted, each lasting $500$ timesteps, with initial parameters evenly spaced across the specified input range. The control and estimation errors for all approaches were averaged and compared against the relative target speed (Figure \ref{fig:fast_test}). The results demonstrate that the proposed approach can effectively track a fast-moving target. While control and estimation errors decrease with lower relative target speeds, performance only significantly drops for the circular case when the target relative speed exceeds $0.7$.

\section{CONCLUSION}

In this work, we presented a bearing-only target tracking and circumnavigation approach utilising a Long Short-Term Memory as the target pose and velocity estimator. Through simulations, we demonstrated the proposed approach's ability to track a fast target moving with a time-varied velocity and withstand input noise. When compared to previously presented circumnavigation and tracking approaches, the presented method displayed lower control and estimation error for the evaluated cases. Extensions of this work could explore multi-agent or multi-target cases. Additionally, tracking controllers that maintain a set distance from the target but do not circumnavigate could be explored.

\bibliographystyle{ieeetr} 
\bibliography{references} 

\end{document}